\pdfoutput=1
\PassOptionsToPackage{table}{xcolor}

\documentclass[11pt]{article}

\usepackage[]{acl}
\usepackage{colortbl}

\usepackage{times}
\usepackage{latexsym}
\usepackage{amsmath}
\usepackage{enumitem} % Add this in your preamble

\definecolor{softgreen}{rgb}{0.5, 0.75, 0.5}
\definecolor{darkgreen}{rgb}{0.0, 0.5, 0.0}

\usepackage[T1]{fontenc}
\usepackage[utf8]{inputenc}
\usepackage{graphicx}
\usepackage{microtype}
\usepackage{multirow}
\usepackage{algorithm}
\usepackage{algorithmicx}
\usepackage{algpseudocode} 
\usepackage{graphicx}
\usepackage{algpseudocode}
\usepackage{bbding}
\usepackage{booktabs}
\usepackage{longtable}

\usepackage{multirow}
\usepackage{algorithm}
\usepackage{algorithmicx}
\usepackage{algpseudocode} 
\usepackage{hyperref}

\usepackage{algpseudocode}
\usepackage{bbding}
\usepackage{booktabs}

\title{Mitigating Training Imbalance in LLM Fine-Tuning   via \\ Selective Parameter Merging}

\author{
    Yiming Ju\textsuperscript{\rm 1}, \ \ 
    Ziyi Ni\textsuperscript{\rm 2},  \  \ 
    Xingrun Xing\textsuperscript{\rm 1, 2},  \  \ 
    Zhixiong Zeng\textsuperscript{\rm 3}, \   \ 
    \textbf{Hanyu Zhao\textsuperscript{\rm 1},} \\ 
    \textbf{Siqi Fan\textsuperscript{\rm 1},} \ \ 
    \textbf{Zheng Zhang\textsuperscript{\rm 1*}}
    \\
    \textsuperscript{\rm 1} Beijing Academy of Artificial Intelligence \\
    \textsuperscript{\rm 2} Institute of Automation, Chinese Academy of Sciences \ \ \ \
    \textsuperscript{\rm 3} Tencent \\
    \texttt{ymju@baai.ac.cn, zhangz.goal@gmail.com} \\
}

\begin{document}
\maketitle
\begin{abstract}

Supervised fine-tuning (SFT) is crucial for adapting Large Language Models (LLMs) to specific tasks. 
In this work, we demonstrate that the order of training data can lead to significant training imbalances, potentially resulting in performance degradation.
Consequently, we propose to mitigate this imbalance by merging SFT models fine-tuned with different data orders, thereby enhancing the overall effectiveness of SFT. 
Additionally, we introduce a novel technique, "parameter-selection merging," which outperforms traditional weighted-average methods on five datasets. 
Further, through analysis and ablation studies, we validate the effectiveness of our method and identify the sources of performance improvements.

\end{abstract}

\section{Introduction}

\renewcommand{\thefootnote}{\fnsymbol{footnote}} % 改变脚注编号样式为符号
\footnotetext[1]{Corresponding Author} % 使用符号脚注
\renewcommand{\thefootnote}{\arabic{footnote}} % 恢复正常编号

Thanks to the substantial expansion of training scale and model size, large language models (LLMs) have achieved significant breakthroughs across a broad spectrum of NLP tasks 
\citep{radford2019language, touvron2023llama}. 
For downstream tasks, supervised fine-tuning (SFT) is a crucial technique for LLMs, enabling the customization of pre-trained models for specialized tasks and domains \citep{dettmers2023qlora, zhao2023survey}.

\begin{figure}[t!]
\centering
\includegraphics[width=1\linewidth]{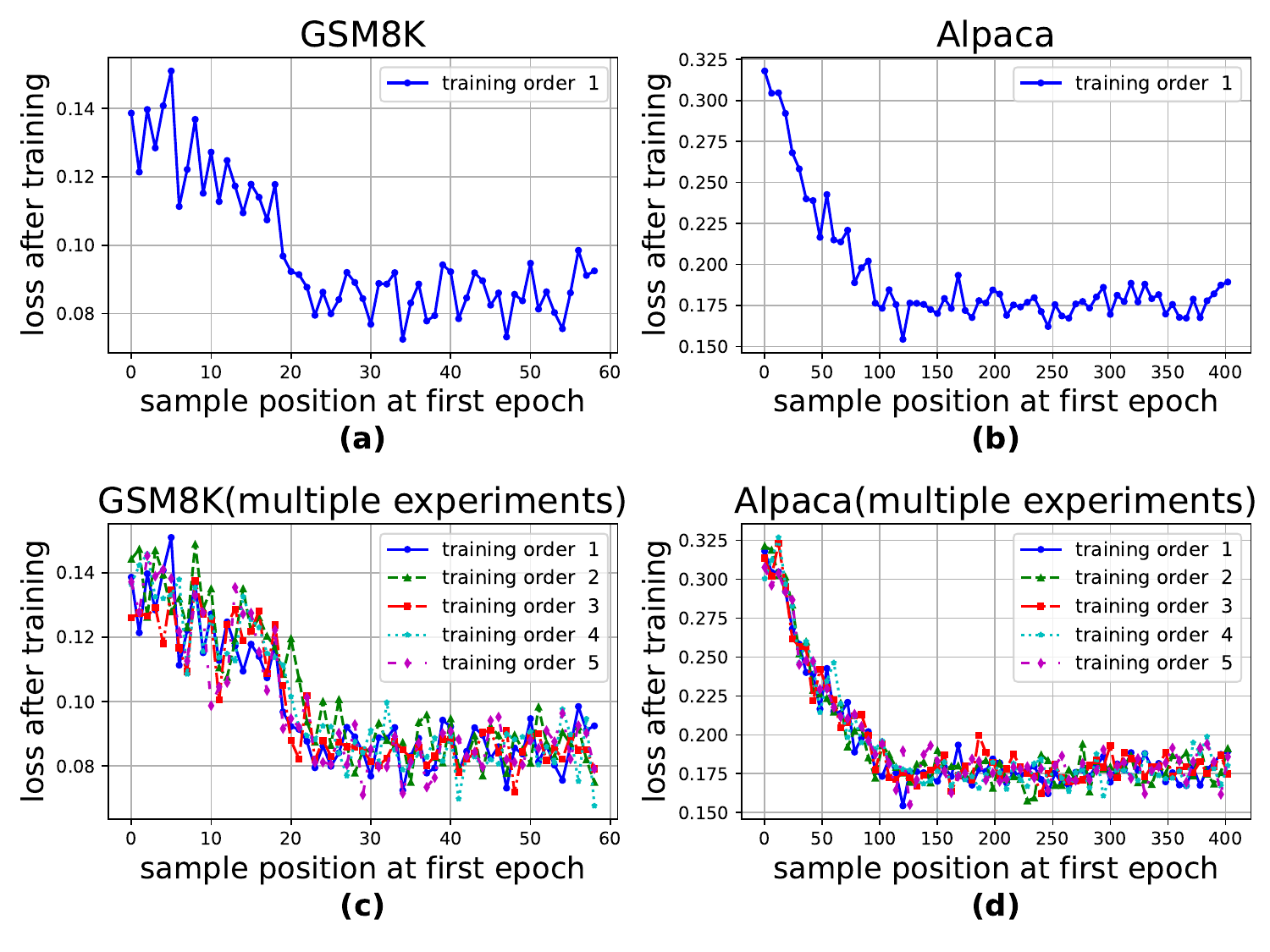}
\caption{ Impact of training sample position at first epoch on final model losses of these samples (after 3 epochs of training). Panels (a) and (b) present the results on the GSM8k and Alpaca tasks, respectively. Panels (c) and (d) show the corresponding results from multiple experiments with different training orders. }
\label{order}	
\end{figure}

The SFT process typically involves a few iterations of training on task-specific data. While existing research generally assumes that the order of training samples has a negligible impact on final model performance, or that sufficient iterations can mitigate any potential effects, our preliminary investigations suggest otherwise. 
We found that the position of SFT training samples significantly affects their final training outcomes.
For instance, Figure \ref{order} (a) and (b) illustrate the relationship between the position of training samples in the first epoch and their losses after three epochs of training.
The figure clearly shows that despite multiple epochs of training, samples introduced earlier consistently exhibit higher final losses. 
Figure \ref{order} (c) and (d) present the results of multiple experiments with different training orders, demonstrating a strong and consistent correlation between the position of training samples and their final losses.\footnote{The experiment was conducted using GSM8K \citep{cobbe2021training} and Stanford Alpaca \citep{taori2023stanford} datasets, with Llama-2-7b \citep{touvron2023llama} as the base model. Each epoch featured a different sample order.}

These findings suggest a notable imbalance in fine-tuning process: samples processed at different positions unevenly influence the learning process, thereby posing a potential risk of skewing the performance of the fine-tuned model.
To mitigate this imbalance, we propose merging multiple SFT models obtained from diverse training data orders through parameter merging technique \citep{matena2022merging}. Moreover, we introduce "\textbf{parameter-selection merging}," a novel parameter merging method that outperforms the traditional weighted-average method. The core contributions of this paper are summarized as follows:

% 1. We identify the training imbalance during model fine-tuning, where the position of training samples significantly affects their final training losses.
    
% 2. We propose to enhance model fine-tuning by merging models trained with different data orders. Moreover, we introduce a novel parameter merging method, "parameter-selection merging."

% 3. Through analysis and ablation studies, we further validate the effectiveness of our method  and demonstrate that the performance improvements are attributed to varied sample positions rather than different batch combinations.
    
\begin{itemize}
% \begin{itemize}[leftmargin=*]

    \item We identify the training imbalance in SFT process, where the position of training samples significantly affects their final training losses.
    
    \item We propose to improve model fine-tuning by merging models trained with different data orders. Moreover, we introduce a novel parameter merging method, "parameter-selection merging."

    \item Through analysis and ablation studies, we further validate the effectiveness of our method and demonstrate the source of improvement.

% and demonstrate that the performance improvements are attributed to varied sample positions rather than different batch combinations.
\end{itemize}

\section{Method}

\subsection{Merge Fine-tuned LLMs with Different Data Order}
In this work, we propose to mitigate training imbalance in LLM fine-tuning by merging models fine-tuned with various data orders. 
As depicted in Figure \ref{merge}, for a given task $t$, the method initiates by fine-tuning a pre-trained LLM multiple times, each with a uniquely ordered data sequence.
Specifically, for various data sequences \(\{s_t^1, s_t^2, \cdots, s_t^k\}\), we obtain a set of SFT models \(\{\boldsymbol{\theta}^{s_t^1}_{SFT}, \boldsymbol{\theta}^{s_t^2}_{SFT}, \cdots, \boldsymbol{\theta}^{s_t^k}_{SFT}\}\). Subsequently, these variously fine-tuned models are integrated into a unified model through parameter merging techniques, yielding an improved SFT model \(\boldsymbol{\theta}_{\text{SFT}}\uparrow\)

\subsection{Parameter-Selection Merging}
\begin{figure}[t!]
\centering
\includegraphics[width=1.0\linewidth]{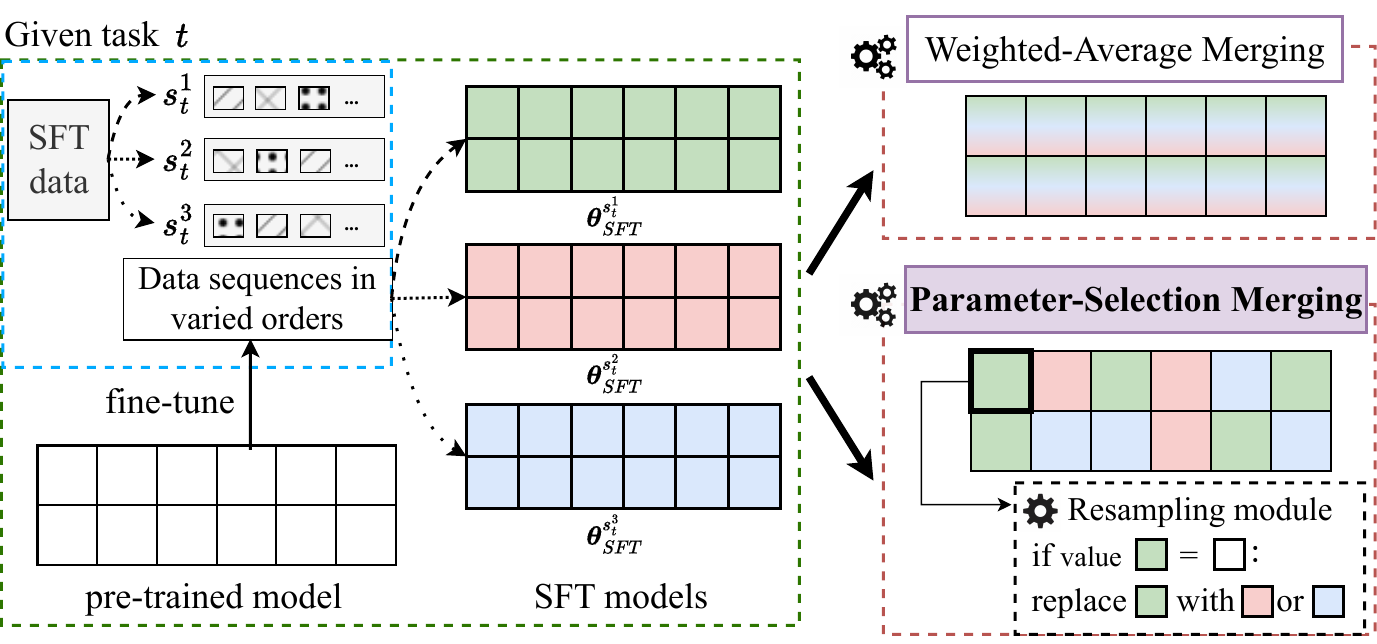}
\caption{Illustration comparing weighted-average method and the proposed parameter-selection method.
Weighted-average merging calculates the weighted sum of all sub-model parameters at each parameter dimension, whereas parameter-selection merging selects parameters from a single sub-model.
In the resampling module, parameters that equal those of the base model are replaced with parameters from alternative models.
}
\label{merge}	
\end{figure}

Existing parameter merging techniques can generally be categorized under "weighted-average merging" approach.
In this work, we introduce a novel parameter merging approach: "\textbf{parameter-selection merging}."  
Figure \ref{merge} shows the comparison of two merging techniques.
Given a set of \(K\) sub-models \(\{\boldsymbol{\theta}_1, \boldsymbol{\theta}_2, \ldots, \boldsymbol{\theta}_K\}\), each model \(\boldsymbol{\theta}_i\) is comprised of parameters \(\theta_{i,1}, \theta_{i,2}, \ldots, \theta_{i,d}\) across \(d\) parameter dimensions.
Weighted-average merging calculates the weighted sum of all sub-model parameters at each parameter dimension, which can be represented by the following formula:
\begin{equation}
 \theta_{\text{merged}, j} = \sum_{i=1}^{K} w_i \theta_{\text{i,j}}, \ \ \forall j \in \{1, \ldots, d\}
\end{equation}
where \(\theta_{i,j}\) is the parameter of the \(i\)-th sub-model in dimension $d$, $w_i$ is the weight applied to  \(\theta_{i,j}\).

Conversely, parameter-selection merging selects a parameter from a single sub-model for each dimension with probbability $p_i$, as represented by the formula:
\begin{equation}
 \theta_{\text{merged}, j} = \theta_{i,j} \text{ with } p_i, \ \ \forall j \in \{1, \ldots, d\}
\end{equation}
where $p_i$ is the probability that \(\theta_{i,j}\) is selected. 
Given that each sub-model in our method is fine-tuned on the same training dataset and thus has nearly identical performance, we assign equal weights and selection probabilities among sub-models, set as:
$
 w_i = \frac{1}{K},  p_i = \frac{1}{K}.
$

% Given that each sub-model in our method is fine-tuned on the same training dataset and thus has nearly identical performance, we assign equal weight \(w_i\) and selection probability \(p_i\) to each sub-model, both set as:
% $
% w_i = \frac{1}{K}, \quad p_i = \frac{1}{K},
% $
% where \(K\) is the total number of sub-models.

% where \(K\) is the total number of sub-models.
% 

% where \(p_i\) is the probability that \(\theta_{i,j}\) is selected. Since each sub-model is fine-tuned on the same dataset and thus has nearly identical performance, we assign equal weight \(w_i\) and selection probability \(p_i\) to each, set as:
% $
% w_i = \frac{1}{K}, \quad p_i = \frac{1}{K},
% $
% with \(K\) representing the total number of sub-models.

\begin{table*}[t!]
\centering
\resizebox{0.95\linewidth}{!}{%
\begin{tabular}{l|cccccc}
\toprule[0.8pt]
\multirow{2}{*}{Method / Task} & AlpacaEval & GSM8K & GSM8K-RFT & MATH & HumanEval & \multirow{2}{*}{Avg $\Delta$} \\
 & win-rate & acc & acc & acc & pass@1 &  \\
\midrule[0.2pt]

single SFT  & 24.25 & 41.29 & 52.74 & 10.36 & 26.82 & - \\

weighted-avg &   24.97{\small(\textcolor{red}{+0.72})} & 44.35{\small(\textcolor{red}{+3.06})} & 53.29{\small(\textcolor{red}{+0.88})} & 11.24{\small(\textcolor{red}{+0.55})} & 26.22{\small(\textcolor{darkgreen}{-0.60})} & + 0.92\\
% \midrule[0.2pt]

\midrule[0.05pt]

param-selection &  25.66{\small(\textcolor{red}{+1.41})} & 44.73{\small(\textcolor{red}{+3.44})} & 53.35{\small(\textcolor{red}{+0.61})} & 11.37{\small(\textcolor{red}{+1.01})} & 27.43{\small(\textcolor{red}{+0.61})} & + 1.42 \\

% \rowcolor[gray]{0.9}
. + resample  &  \textbf{25.91}{\small(\textcolor{red}{+1.66})} & \textbf{45.26}{\small(\textcolor{red}{+3.97})}  & \textbf{54.32}{\small(\textcolor{red}{+1.58})} & \textbf{12.00}{\small(\textcolor{red}{+1.64})} & \textbf{28.05}{\small(\textcolor{red}{+1.23})} & \textbf{+ 2.02} \\

\bottomrule[0.8pt]
\end{tabular}%
}
\caption{Performance comparison of weighted-average and parameter-selection merging based on Llama-2-7b. "weighted-avg" means weighted-average and "param-selection" means parameter-selection merging method. }
\label{re2}
\end{table*}

\subsection{Resample Strategy}
\noindent \textbf{Task Vectors}.
Let $\boldsymbol{\theta}_{\text{pre}}$ represent the pre-trained model's weights and $\boldsymbol{\theta}_{\text{SFT}}$ denote the SFT model 's weights. The task vector $\tau$ is defined to capture task-specific adaptations, calculated as: $\boldsymbol{\tau} = \boldsymbol{\theta}_{\text{SFT}} - \boldsymbol{\theta}_{\text{pre}}$ \citep{ilharco2022editing}.

Guided by the intention to maximize the impact of  task vectors, we introduce a resampling method within the parameter-selection merging framework to further improve task performance. 
$\tau_{i,j}$ represents the task vector of the $i$-th sub-model at parameter dimension $j$.
As depicted in Figure \ref{merge}, if $\tau_{i,j} = 0$, indicating that no parameter change occurred after fine-tuning, a new parameter is resampled from the pool of all sub-models.\footnote{This strategy enables parallel tensor operations by including all sub-models in resampling, not just the remaining ones.} This procedure can be iterated \(n\) times, where \(n\) is a predefined hyperparameter, as formalized below:
\begin{equation}
\theta_{\text{merged}, j}^{(n)} = 
\begin{cases} 
\theta_{i,j} & \text{if } \tau_{i,j} \neq 0 \text{ or } n = 0, \\
\theta_{\text{merged},j}^{(n-1)} & \text{others},
\end{cases}
\end{equation}
Specifically, $\theta_{\text{merged},j}^{\text{ } 0}$ equals parameter-selection method without the resampling module.

\section{Experiments}

This section presents the experimental results. Detailed descriptions of the datasets and evaluation metrics employed are provided in the Appendix, under Section \ref{Experimental Settings}.

\subsection{Experimental Results}
~\\
\noindent
\textbf{Main Experiments.}
We conducted experiments on three mainstream LLM tasks: instruction-following, mathematical reasoning, and code-generating. Llama-2-7b \citep{touvron2023llama} was used as the base model. As shown in Table \ref{re2}, the merged models exhibit performance improvements compared to single SFT models.
Furthermore, as indicated in Table \ref{re2}, the proposed parameter-selection method outperforms the weighted-average approach, achieving consistent performance improvements. Moreover, incorporating a resampling module further enhances the performance of the parameter-selection method, yielding an average improvement of \textbf{2.02} percentage points across all datasets. These results affirm the effectiveness of our proposed method in improving LLM fine-tuning performance.

\begin{table}[t!]
\centering
\resizebox{\linewidth}{!}{%
\setlength{\tabcolsep}{2.5pt} % 减少列间距
\begin{tabular}{ll|cccc}
\toprule[0.8pt]
 \multirow{2}{*}{Model}  &  \multirow{2}{*}{Method}  & SST-2 & MNLI & SQuAD & \multirow{2}{*}{Avg $\Delta$} \\
 &   & acc & acc & EM &  \\
\midrule[0.4pt]
% \hline
\multirow{2}{*}{BERT-base} &  SFT & 91.93 & 83.99 & 81.07 & \multirow{2}{*}{\normalsize\textcolor{black}{+ 0.75}} \\
& merged & 92.33 \small(\textcolor{red}{+0.40}) & {84.47}\small(\textcolor{red}{+0.48}) & {82.44}\small(\textcolor{red}{+1.37})  \\
\midrule[0.2pt]
\multirow{2}{*}{BERT-large} &  SFT & 93.44 & 86.42 & 84.15 & \multirow{2}{*}{\normalsize\textcolor{black}{+ 0.94}} \\
& merged  & 
{94.38}{\small(\textcolor{red}{+0.94})} 
& 86.71{\small(\textcolor{red}{+0.29})} & {85.73}{\small(\textcolor{red}{+1.58}) }
\\
\midrule[0.2pt]
\multirow{2}{*}{TinyLlama} &  SFT & 94.81 & 85.46 & 80.53 & \multirow{2}{*}{\normalsize\textcolor{black}{+ 1.62}} \\
& merged  
& {95.91}{\small(\textcolor{red}{+1.10})} & {86.93}{\small(\textcolor{red}{+1.47})} & {82.82}{\small(\textcolor{red}{+2.29})}\\
\midrule[0.2pt]
\multirow{2}{*}{Llama-2-7b} &  SFT & 95.09 & 88.84 & 84.53 & \multirow{2}{*}{\normalsize\textcolor{black}{+ 2.11}} \\
 & merged 
 & {96.97}{\small(\textcolor{red}{+1.88})} & {90.64}{\small(\textcolor{red}{+1.80})} & {87.18}{\small(\textcolor{red}{+2.65})} \\
\bottomrule[0.8pt]
\end{tabular}
}
\caption{Performance comparison between single SFT model and merged models across pre-trained models with various model sizes.}
\label{re1}
\end{table}

~\\
\noindent
\textbf{Experiments Across Different Model Sizes.}
We conducted experiments using different pre-trained models with various model sizes: BERT-base (0.11b)\footnote{(0.11b) refers to the model having 0.11 billion parameters.}, BERT-large (0.34b) \citep{kenton2019bert}, TinyLlama (1.1b) \citep{zhang2024TinyLlama}, and Llama-2-7b (7b), employing parameter-selection as merging method.
% \footnote{Experiments were conducted on traditional tasks rather than on LLM tasks due to the limited capabilities of smaller-sized models.}
Experiments were conducted on traditional tasks rather than on LLM tasks due to the limited capabilities of smaller-sized models.
As shown in Table \ref{re1}, the merged models outperform their single SFT counterparts consistently. These experimental outcomes further demonstrate the effectiveness of merging SFT models with different training orders in improving fine-tuning performance. Furthermore, as detailed in Table \ref{re1}, models with larger parameter sizes exhibit more pronounced average improvements, suggesting our method's potential applicability in LLM contexts.

% \begin{table*}[t!]
% \centering
% \resizebox{0.9\linewidth}{!}{%
% \begin{tabular}{lccccccc}
% \toprule[0.8pt]
% \multirow{2}{*}{Method}  & AG News & Hellaswag & MNLI & MRPC & SST-2 & Winogrande & \multirow{2}{*}{Avg $\Delta$}\\
% & acc & acc & acc & acc & acc & acc & \\
% \midrule[0.2pt]
% \multicolumn{8}{c}{\textit{Single-Task \ Model}} \\
% \midrule[0.2pt]
% single SFT  & 94.42   & 77.20 & 87.90 & 85.78 & 95.53 & 75.45 & -\\
% \midrule[0.2pt]
% \multicolumn{8}{c}{\textit{Multi-Task \ Models}} \\
% \midrule[0.2pt]
% weighted-avg &  74.01 & 74.10 & 61.15 & 71.32 & 90.37 & 70.17 & - 12.53 \\

% % \midrule[0.2pt]

% param-selection & 77.03 & \textbf{74.13} & 64.77 & 67.16 & 92.66 & 70.40 & - 11.67 \\
% \rowcolor[gray]{0.9}
% . + resample  & \textbf{81.28} & 74.12 & \textbf{64.45} & \textbf{72.55} & \textbf{95.30} & \textbf{70.56} & \textbf{- \ \ 9.67} \\

% \bottomrule[0.8pt]
% \end{tabular}
% }
% \caption{Performance comparison in multi-task merging contexts for traditional tasks. ". + resample" refers to the addition of the resampling module to our parameter-selection method.}
% \label{re3}
% \end{table*}

\begin{table*}[t!]
\centering
\resizebox{\linewidth}{!}{%
\setlength{\tabcolsep}{2.5pt} % Decrease column spacing
\begin{tabular}{l|ccccc|cccccccc}
\toprule[0.8pt]
 % \multirow3{*}{} 
 & \multicolumn{5}{c|}{LLM Tasks}  & \multicolumn{7}{c}{Traditional Tasks} \\
% \cline{3-5}  \cline{8-12}
% \midrule[0.2pt]
\cmidrule(lr){2-13}
% \cline{2-13}

Method / Task   & \small AlpacaEval & \small GSM8K & \small MATH & \small HumanEval & \multirow{2}{*}{\small Avg $\Delta$} & \small AG News &  \small Hellaswag &  \small MNLI &  \small MRPC &  \small SST-2 &  \small Winogrande & \small \multirow{2}{*}{Avg $\Delta$}\\
 & win-rate & acc & acc & pass@1 & & acc & acc & acc & acc & acc & acc & \\
\midrule[0.2pt]
 % \cmidrule(lr){1-13}

% & \multicolumn{4}{c}{\textit{Single-Task \ Model}} & & \multicolumn{7}{c}{\textit{Single-Task \ Model}} & \\
% % \cmidrule(lr){2-5}

% \midrule[0.2pt]
single SFT & 89.29 & 63.76 & 14.26 &  23.78 & - & 94.42   & 77.20 & 87.90 & 85.78 & 95.53 & 75.45 & -\\
\midrule[0.2pt]
 % \cmidrule(lr){1-13}

% & \multicolumn{5}{c|}{\textit{Multi-Task \ Models}} & \multicolumn{7}{c}{\textit{Multi-Task \ Models}} \\
% % \midrule[0.2pt]
% % \cmidrule(lr){2-5}

weighted-avg & 72.29 & 58.38 & 9.90 &  18.90 & - 7.91 &  74.01 & 74.10 & 61.15 & 71.32 & 90.37 & 70.17 & - 12.53 \\
param-selection & 72.08 & 57.01 & 10.1 &  14.64 & - 9.32 & 77.03 & \textbf{74.13} & 64.77 & 67.16 & 92.66 & 70.40 & - 11.67 \\
% \rowcolor[gray]{0.9}
. + resample & \textbf{78.70} & \textbf{61.71} & \textbf{11.7} & \textbf{26.22} & \textbf{- 3.19} & \textbf{81.28} & 74.12 & \textbf{64.45} & \textbf{72.55} & \textbf{95.30} & \textbf{70.56} & \textbf{- \ \ 9.67} \\
\bottomrule[0.8pt]
\end{tabular}%
}
\caption{
% Performance comparison in multi-task merging contexts.
% The "single SFT" represents a single-task model, showing results for individual tasks, whereas the other entries are multi-task models, showing results for handling multiple tasks simultaneously.
Performance comparison in multi-task merging contexts.
The "single SFT" represents a single-task model, showing results for individual tasks, whereas the other entries are multi-task models, showing results for handling multiple tasks simultaneously.
Avg $\Delta$ shows the average performance loss across all tasks after merging these single-task models into a single multi-task model.
}
\label{re4}
\end{table*}

~\\
\noindent
\textbf{Experiments in Multi-Task Merging Contexts.}
We conducted experiments in multi-task merging contexts to validate the effectiveness of parameter-selection.\footnote{Multi-task merging aims to combine single-task models into one multi-task model capable of handling several tasks simultaneously, with minimal performance loss in single-task capabilities.}
The results are presented in Table \ref{re4}, which show the performance on four LLM tasks and six traditional tasks.\footnote{Due to significant performance degradation for LLM tasks, 13b models were chosen instead of 7b. Specifically, we used WizardLM-13B \citep{xu2023wizardlm} for instruction-following, WizardMath-13B \citep{luo2023wizardmath} for mathematical reasoning, and Llama-2-13b-code-alpaca \citep{chaudhary2023code} for code-generating.}
As demonstrated by the results, the parameter-selection method significantly outperforms the weighted-average method, achieving an increase of \textbf{4.72} and \textbf{2.86} percentage points in performance retention, respectively. These improvements highlight the efficacy of the proposed parameter-selection method.

% In multi-task merging contexts, we conduct experiments on six traditional tasks using Llama-2-7b as the base model. The results are presented in Table \ref{re3}. Consistent with the results for LLM tasks, the parameter-selection method outperforms the average-based method as well, achieving \textbf{2.86} more percentage points in performance retention.

\subsection{Analysis and Ablation Studies}
This section presents the analysis and ablation studies conducted on the GSM8K and Alpaca tasks.

\begin{figure}[t!]
\centering
\includegraphics[width=1\linewidth]{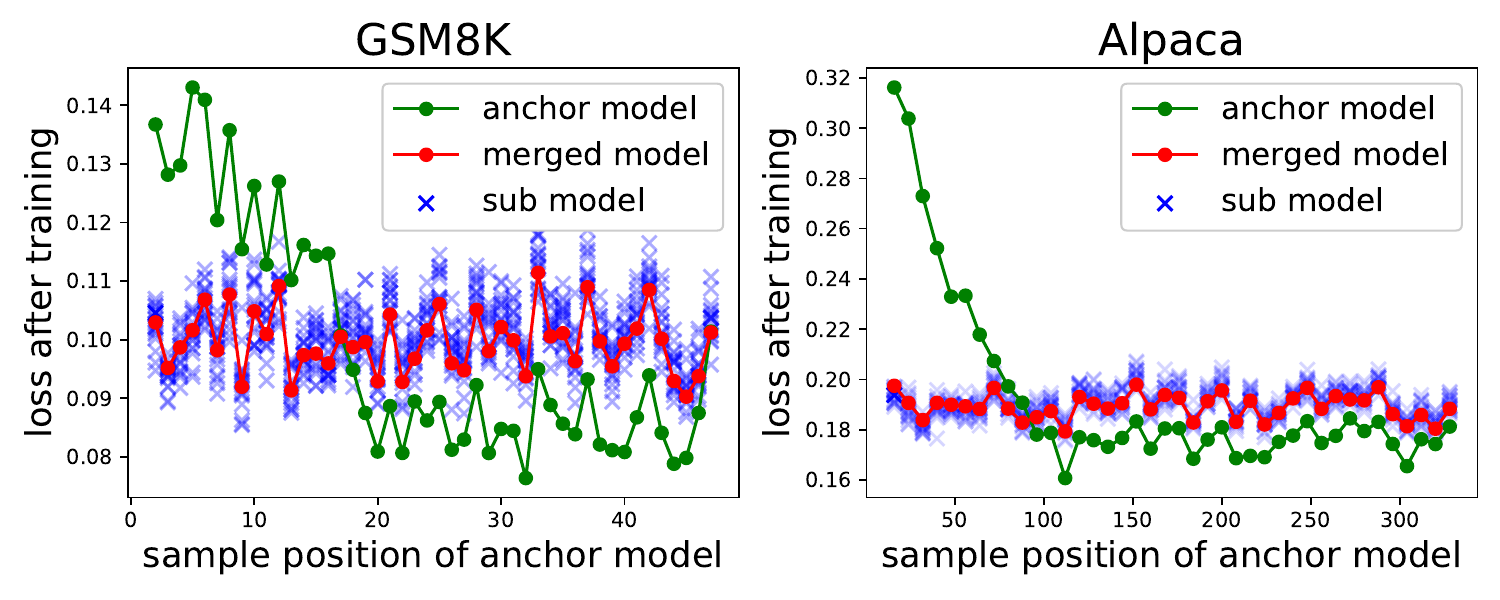}
\caption{Comparison of training losses across different models, with the first epoch sample position of the anchor model as the x-axis. Green lines represent final training losses of the anchor model; blue `x' markers indicate losses of SFT models trained with various data order; red dots show losses of the merged model.}
\label{train loss}	
\end{figure}

~\\
\noindent
\textbf{Traning Set Loss Analysis.}
We investigate whether the merged models can alleviate the training imbalance problem previously identified.
We selected one SFT model as the "\emph{anchor model}". Based on positions during the first epoch training of the anchor model, we divided training samples into multiple segments.
Figure \ref{train loss} shows the final training loss of these sample segments.
As shown in Figure \ref{train loss}, compared to the anchor model, the losses of the merged model are situated between those of sub-models, showing no clear correlation with the data position. 
This result indicates that merging models with various data orders can diminish the influence of the data order from a single model, such as the anchor model.

\begin{figure}[t!]
\centering
\includegraphics[width=1\linewidth]{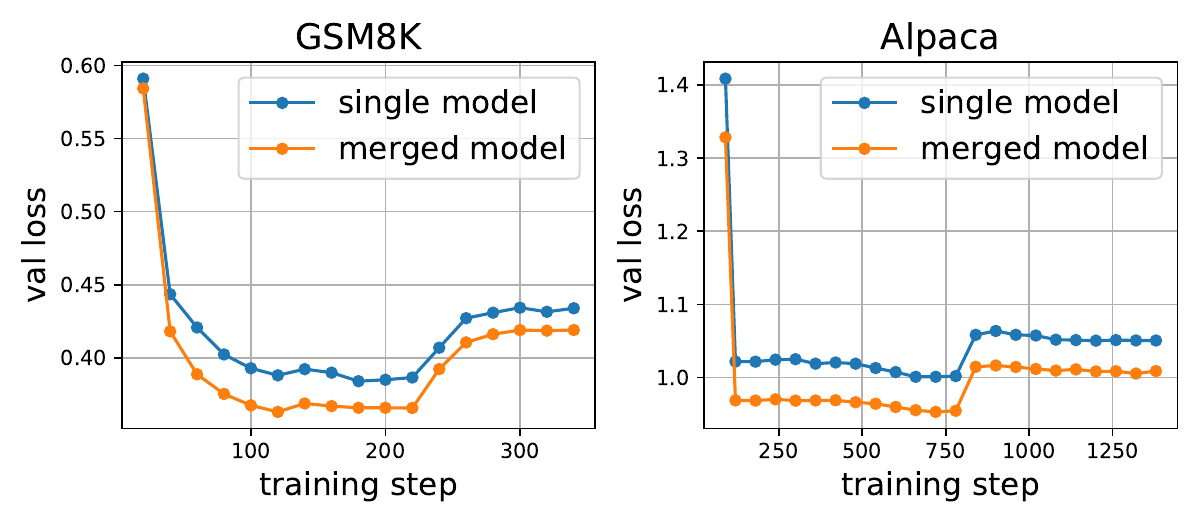}
\caption{Comparison of validation loss between single and merged SFT models at various training steps.}
\label{loss}
\end{figure}

~\\
\noindent
\textbf{Validation Set Loss Analysis.} We analyzed the validation set losses of the single SFT model and the merged model at various training steps. As shown in Figure \ref{loss}, at all training steps, the merged models exhibited lower validation losses compared to those of single SFT models. 
This result demonstrates that the merged model exhibits lower losses on unseen samples, which aligns with the performance enhancements previously observed.

\begin{table}[t!]
\centering
\resizebox{0.98\linewidth}{!}{%
\setlength{\tabcolsep}{2.5pt} % Decrease column spacing
\begin{tabular}{lcc}
\toprule[0.8pt]
\multirow{2}{*}{Method} & \small GSM8K  &  \small AlpacaEval \\
 &acc   &  win-rate  \\
\midrule[0.2pt]
singel SFT & 41.29 & 24.25 \\
param-selection + resample  & 45.26 & 25.91 \\
param-selection + resample  (fix-batch) & 45.51 & 25.83 \\
\bottomrule[0.8pt]
\end{tabular}%
}
\caption{Performance comparison of standard merged models and models with fixed intra-batch combinations.
}
\label{fix_batch}
\end{table}

% ~\\
% \noindent
% \textbf{Determining the Source of Improvement: Sample Position or Batch Diversity.}
% Altering the order of training data not only changes the position of samples but also modifies the combinations of samples within each batch.
% This raises the question: Does performance improvement result from varied sample positions or diversity in sample combinations?
% To address this, we conducted ablation experiments by merging models with fixed intra-batch sample combinations while varying batch positions.
% As shown in Table \ref{fix_batch}, models with fixed intra-batch combinations achieved similar performance to those with variable combinations, indicating that performance gains are primarily due to changes in sample positions rather than the diversity in intra-batch combinations.

~\\
\noindent
\textbf{Determining the Source of Improvement: Sample Position or Batch Diversity.}
Altering the order of training data not only changes the position of samples but also modifies the sample combination within each batch.
This raises the question: Do performance improvements result from varied sample positions or from diversity in sample combinations?
To address this, we conducted ablation experiments by merging models with fixed intra-batch sample combinations while varying batch positions.
As shown in Table \ref{fix_batch}, models with fixed intra-batch combinations achieved similar performance to those with variable combinations, indicating that performance gains are primarily due to changes in sample positions rather than to diversity in intra-batch combinations.

\section{Conclusion}

This study reveals the overlooked negative impact of training data order on LLM fine-tuning, which can result in significant training imbalances, and proposes mitigating these imbalances by merging models fine-tuned with diverse data orders.
Furthermore, it introduces a novel and effective parameter merging technique, "parameter-selection merging."
The efficacy of parameter-selection method suggests that it is not necessary to incorporate information from all sub-models at each parameter dimension in parameter merging.
This discovery broadens the research landscape for parameter merging, opening up new avenues for future investigations.

\section*{Limitations}
This study has several primary limitations that remain unexplored:
\begin{itemize}
    \item While our method improves LLM fine-tuning without adding deployment and inference costs, it requires additional computation to fine-tune multiple sub-models.
    \item Although models with larger parameter sizes show more pronounced average improvements, as demonstrated in Table \ref{re1}, suggesting the method's potential in LLM contexts, our experiments were primarily conducted with 7b models due to computational resource constraints. Future studies are needed to evaluate the scalability of our methods with larger models.
    \item The study introduces the novel parameter-selection merging technique, which outperforms the traditional weighted-average approach. However, many model merging studies in multi-task scenarios rely on a weighted-average formula. It remains to be explored whether replacing the weighted-average with parameter-selection can improve these existing methods in multi-task scenarios.
\end{itemize}

\section*{Acknowledgements}
% This work is supported by National Key R\&D Program
% of China 2022ZD0116301.
This work is supported by the National Science and Technology Major Project (2022ZD0116301) and the National Science Foundation of China under grant No.62206150.

% 1.  Although parameter merging method can enhance LLM fine-tuning without adding deployment and inference costs, it need more computation to fine-tune multiple sub-models.
% 2. The study predominantly centers on sub-models derived from varying training data sequences.
% Broader range of configurations, such as merging models trained with different learning rates, training steps, or batch sizes are unexplored. This limitation points to potential avenues for future research, where these variables can be systematically explored to understand their impact.
% 3. The study improves single-task fine-tuning performance through model merging and introduces parameter-selection merging technique.
% While previous studies mainly concentrated on multi-task model merging. 
% The opportunity to apply parameter-selection in lieu of weighted averaging for multi-task scenarios presents an intriguing prospect for future work. 
% Specifically, the traditional weighted average's weights (\(w_i\)) could be substituted with the selection probabilities (\(p_i\)) from parameter-selection merging to replicate and compare against existing methodologies in multi-task contexts.

\bibliography{anthology,custom}

\appendix

\begin{table*}[t!]
\centering
\resizebox{\linewidth}{!}{%
\setlength{\tabcolsep}{4pt} % 减少列间距
\begin{tabular}{l|ccc|ccccccc}
\toprule[0.8pt]
Model  & \multicolumn{3}{c|}{BERT-base \& BERT-large} & \multicolumn{7}{c}{TinyLlama \& Llama-2-7b} \\
Dataset  & SST-2 & MNLI & SQuAD & SST-2 & MNLI & SQuAD & AG News & Hellaswag & MRPC & Winogrande \\
\midrule[0.2pt]
max seq-length & 512  & 512 & 512  & 800 & 800 & 800 & 800 & 800 & 800 & 800 \\
learning rate &2e-5  & 2e-5  & 3e-5 & 2e-5 & 2e-5 &2e-5 & 2e-5 & 2e-5 & 2e-5 & 2e-5 \\
batch size & 32 & 32 & 12 & 128 & 128 & 128 & 128 & 128 & 128 & 128\\
\bottomrule[0.8pt]
\end{tabular}%
}
\caption{Hyperparameters for training models on traditional tasks.}
\label{param1}
\end{table*}

\begin{table*}[t!]
\centering
\begin{tabular}{l|ccccccc}
\toprule[0.8pt]
Dataset  & AlpacaEval & GSM8K & GSM8K-RFT & MATH & HumanEval \\
\midrule[0.2pt]
max seq-length &1200 & 800 & 800 & 800 & 1200 \\
learning rate &2e-5 & 2e-5 & 2e-5 & 2e-5 & 2e-5 \\
batch size &128 & 64 & 64 & 64 & 128 \\
max epoch &3 & 3 & 3 & 3 &  3 \\
$n$ & 1 & 1 & 4 & 1 & 4 \\
\bottomrule[0.8pt]
\end{tabular}
\caption{Hyperparameters for training Llama-2-7b on LLM tasks.}
\label{param2}
\end{table*}

\section{Related Work}
\subsection{Parameter Merging in Multi-Task Scenario}

Parameter merging, defined as combining multiple models within the parameter space \citep{matena2022merging}, primarily focuses on integrating SFT models for different tasks into one capable of addressing all associated sub-tasks (multi-task scenario).
Numerous related studies have been conducted in this field. 
For example, 
\citet{wortsman2022model} and \citet{jin2022dataless} employed linear matrix transformation for task adaptability;
\citet{yadav2023ties} addressed the issue of sign conflicts across different sub-tasks;
Similarly,
\citet{yu2023language} mitigated task conflict by
partially removing task-specific parameters;
Moreover, \citet{xiao2023lm} aimed to maximally preserve the performance of one primary task among all tasks;
Furthermore, \citet{huang2023lorahub} investigated the composability of LoRA  \citep{hu2021lora} for enhancing cross-task generalization.

\subsection{Parameter Merging in Single-Task Scenario}

Compared to merging models from multiple tasks, which often leads to performance degradation on individual tasks, the potential of utilizing the parameter merging technique to improve single-task LLMs has not yet received much attention.
While some studies, such as \citet{wortsman2022model}, have explored merging models fine-tuned with different settings, these experiments were predominantly conducted on comparatively smaller models like BERT and achieved only modest improvements.

\section{Detailed Experimental Settings}\label{Experimental Settings}

\subsection{Datasets}\label{Datasets}
Datasets employed in our experiments are categorized into two groups: LLM tasks and traditional NLP tasks.

~\\
\noindent
\textbf{LLM Tasks:}
\begin{itemize}[leftmargin=*]

% \begin{itemize}
    \item \textbf{Instruction-following:} Stanford Alpaca \citep{taori2023stanford}
    \item \textbf{Mathematical Reasoning:} GSM8K \citep{cobbe2021training}, GSM8K-RFT \citep{yuan2023scaling}, MATH \citep{hendrycks2021measuring}
    \item \textbf{Code-generating:} Evol-instruction-66k, obtained from \href{https://huggingface.co/datasets/codefuse-ai/Evol-instruction-66k}{Hugging Face Datasets}
\end{itemize}
~\\
\noindent
\textbf{Traditional NLP Tasks:}
\begin{itemize}[leftmargin=*]
    \item SST-2 \citep{xu2023wizardlm}
    \item MNLI \citep{williams2017broad}
    \item SQuAD \citep{rajpurkar2016squad}
    \item AG News \citep{zhang2015character}
    \item Hellaswag \citep{zellers2019hellaswag}
    \item MRPC \citep{dolan2005automatically}
    \item Winogrande \citep{sakaguchi2020winogrande}
\end{itemize}
For traditional tasks, experiments involving decoder-based models utilized the version collected by \citet{cheng2023uprise, wang2023learning}.
For the MATH dataset, an augmented version \citep{yu2023metamath} is employed, with data originally sourced from GSM8K excluded. The Evol-instruction-66k dataset is obtained from the Hugging Face library (\href{https://huggingface.co/datasets/codefuse-ai/Evol-instruction-66k}{https://huggingface.co/datasets/codefuse-ai/Evol-instruction-66k}).

\subsection{Evaluation Metrics}\label{Evaluation Metrics}
We employ AlpacaEval \citep{li2023alpacaeval} to evaluate models fine-tuned on Stanford Alpaca dataset, using win-rate as the evaluation metric and GPT-4 as the annotator.
We employ HumanEval \citep{chen2021evaluating} to evaluate models fine-tuned on Evol-instruction-66k dataset, using pass@1 as the evaluation metric. 
For the SQuAD dataset, Exact Match (EM) is utilized as the evaluation metric. Accuracy (acc) is used as the evaluation metric for all other tasks.

% \begin{table*}[t!]
% \centering
% \begin{tabular}{lccccccc}
% \toprule[0.8pt]
% \multirow{2}{*}{Method}  & AG News & Hellaswag & MNLI & MRPC & SST-2 & Winogrande & \multirow{2}{*}{Avg $\Delta$}\\
% & acc & acc & acc & acc & acc & acc & \\
% \midrule[0.2pt]
% \multicolumn{8}{c}{\textit{Single-Task \ Model}} \\
% \midrule[0.2pt]
% single SFT  & 94.42   & 77.20 & 87.90 & 85.78 & 95.53 & 75.45 & -\\
% \midrule[0.2pt]
% \multicolumn{8}{c}{\textit{Multi-Task \ Models}} \\
% \midrule[0.2pt]
% weighted-avg &  74.01 & 74.10 & 61.15 & 71.32 & 90.37 & 70.17 & - 12.53 \\

% % \midrule[0.2pt]

% param-selection & 77.03 & \textbf{74.13} & 64.77 & 67.16 & 92.66 & 70.40 & - 11.67 \\
% \rowcolor[gray]{0.9}
% . + resample  & \textbf{81.28} & 74.12 & \textbf{64.45} & \textbf{72.55} & \textbf{95.30} & \textbf{70.56} & \textbf{- \ \ 9.67} \\

% \bottomrule[0.8pt]
% \end{tabular}

% \caption{Performance comparison in multi-task merging contexts for traditional tasks. ". + resample" refers to the addition of the resampling module to our parameter-selection method.}
% \label{re3}
% \end{table*}

\subsection{Basic Settings}
For single SFT models, we report the average results across all sub-models. For parameter-selection merging models, we conduct five experiments with different random seeds and report the average outcomes. For decoder-based models, the temperature is set to 0.0 for greedy decoding. Training of LLMs was conducted using mixed precision BF16. All experiments were conducted on 8 NVIDIA Tesla A800 GPUs.

\subsection{Hyperparameters}
For the parameter merging method, the number of sub-models \( K \) is a necessary hyperparameter. Based on the selection range of 1-50 suggested by \citet{wortsman2022model}, we use \( K = 20 \), a relatively moderate value for all datasets (15 datasets in total). The search space for resampling times \( n \) includes \{1, 2, 3, 4\}.
In our experiments, the maximum number of epochs was set to 3, with model states saved at the end of each epoch. 
The hyperparameters used for fine-tuning are detailed in Tables \ref{param1} and \ref{param2}.

\section{Computational Complexity of Merging Process}
The parameter selection and weighted-average merging processes can be efficiently managed on a CPU with rapid execution times. For instance, merging 10 Llama-2-7b models on a single CPU typically takes about 1 minute. The resampling process, meanwhile, requires time proportional to the number of resampling iterations \( n \), with each iteration approximately taking about 0.1 minute.

% \section{Experiments in Multi-Task Merging Contexts for Traditional Tasks}

% In multi-task merging contexts, we conduct experiments on six traditional tasks using Llama-2-7b as the base model. The results are presented in Table \ref{re3}. Consistent with the results for LLM tasks, the parameter-selection method outperforms the average-based method as well, achieving \textbf{2.86} more percentage points in performance retention.

\end{document}